# Exact Structure Discovery in Bayesian Networks with Less Space

**Pekka Parviainen** and **Mikko Koivisto**
Helsinki Institute for Information Technology HIIT, Department of Computer Science
University of Helsinki, Finland
{pekka.parviainen, mikko.koivisto}@cs.helsinki.fi

## Abstract

The fastest known exact algorithms for score-based structure discovery in Bayesian networks on $n$ nodes run in time and space $2^n n^{O(1)}$. The usage of these algorithms is limited to networks on at most around 25 nodes mainly due to the space requirement. Here, we study space–time tradeoffs for finding an optimal network structure. When little space is available, we apply the Gurevich–Shelah recurrence—originally proposed for the Hamiltonian path problem—and obtain time $2^{2n-s} n^{O(1)}$ in space $2^s n^{O(1)}$ for any $s = n/2, n/4, n/8, \ldots$; we assume the indegree of each node is bounded by a constant. For the more practical setting with moderate amounts of space, we present a novel scheme. It yields running time $2^n (3/2)^p n^{O(1)}$ in space $2^n (3/4)^p n^{O(1)}$ for any $p = 0, 1, \ldots, n/2$; these bounds hold as long as the indegrees are at most $0.238n$. Furthermore, the latter scheme allows easy and efficient parallelization beyond previous algorithms. We also explore empirically the potential of the presented techniques.

## 1 INTRODUCTION

There has been relatively recent interest in devising exact algorithms for score-based structure learning in Bayesian networks (Eaton and Murphy, 2007; Koivisto and Sood, 2004; Koivisto, 2006; Perrier et al., 2008; Silander and Myllymäki, 2006; Singh and Moore, 2005). The research is motivated not only by theoretical curiosity but also by applications: an exact algorithm—that is guaranteed to produce an optimal solution or the exact quantity of interest—allows the user to concentrate on modelling issues and direct interpretation of the learning results, with no (extra) uncertainty on the quality of the algorithm's output *per se*.

The fastest known exact algorithms for Bayesian networks on $n$ nodes (i.e, attributes or variables) compute and store intermediate results for all the possible $2^n$ node subsets, running in time and space $2^n n^{O(1)}$, assuming the score obeys certain usual modularity properties (Koivisto and Sood, 2004; Ott and Miyano, 2003; Silander and Myllymäki, 2006; Singh and Moore, 2005). While both the time and the space requirement become soon prohibitive as $n$ gets larger, it is particularly the space requirement that determines the feasibility limit in practice. Indeed, on typical modern desktop computers with a few gigabytes of memory the algorithms can handle networks on up to around 25 nodes before they run out of space, the running time being still in only some minutes or hours. The current record of 29 nodes was achieved by the streamlined Silander–Myllymäki implementation (Silander and Myllymäki, 2006) using nearly 100 gigabytes of hard disk. For making exact algorithms practically feasible in larger networks, reducing the space complexity is the main concern.

To understand the nature of the algorithmic challenges at hand, it is useful to think the structure discovery problem as a "permutation problem": one seeks a linear order of the $n$ nodes that maximizes a sum of local scores, one per node; the local score for a node only depends on the set of nodes that precede it in the order (the potential endpoints of the incoming arcs). Note that here we restrict ourselves to the problem of finding an optimal network structure. For a number of similar permutation problems—like the Travelling Salesman problem, the Feedback Arcset problem, and Treewidth, to name a few—dynamic programming algorithms running in time and space $2^n n^{O(1)}$ have been known for decades (Bellman, 1962; Held and Karp, 1962; Lawler, 1964; Arnborg et al., 1987), with only negligible progress since; more recent work (Bodlaender et al., 2006) has shown that if only polynomial space is allowed, then many permutation problems can be solved in time $4^n n^{O(1)}$ using a divide and conquer technique we outline in the next two paragraphs. We are not aware of any previous work on interpolating between the two extremes of space complexity.

Perhaps the most straightforward approach to solve a permutation problem in space less than $2^n n^{O(1)}$ is to divide the node set into two buckets: $N_0$ containing the first $s$ nodes in the order, and $N_1$ containing the remaining $n-s$ nodes. Both $N_0$ and $N_1$ specify subproblems that can be solved by an exact algorithm in time and space $2^s n^{O(1)}$ and $2^{n-s} n^{O(1)}$, respectively; here we assume the indegrees of the network, i.e. the number of parents per node, are at most some constant. Assuming w.l.o.g. that $s \geq n/2$, the algorithm that tries out all possible partitions $\{N_0, N_1\}$ runs in time $\binom{n}{s} 2^s n^{O(1)}$ and space $2^s n^{O(1)}$. For instance, putting $s = 4/5n$ yields time $O(2.872^n)$ and space $O(1.742^n)$. This simple scheme is the starting point of the present study of time–space tradeoffs for exact structure discovery in Bayesian networks, in two respects.

First (in Section 3), we notice that the above scheme with balanced partitioning (into about equally sized parts) and recursive application yields time $2^{2n-s} n^{O(1)}$ in space $2^s n^{O(1)}$ for any $s = n/2, n/4, n/8, \ldots$. In particular, we get a polynomial-space algorithm with running time $4^n n^{O(1)}$. This divide and conquer technique is known as the Gurevich–Shelah recurrence, originally presented for the Hamiltonian path problem (Gurevich and Shelah, 1987) and later applied also elsewhere (Björklund and Husfeldt, 2008; Bodlaender et al., 2006). Unfortunately, the Gurevich–Shelah recurrence falls short when one is allowed to use more space, say $2^{4n/5} n^{O(1)}$.

Our second, main contribution (in Section 4) addresses the practically more relevant range where the Gurevich–Shelah recurrence does not apply. By contrast, the above outlined simple scheme (unbalanced and without recursion) applies, but—perhaps somewhat surprisingly—turns out to be suboptimal. Indeed, we present a completely different approach, where the idea is to cover the linear orders by a class of suitably specified partial orders, namely, partial orders consisting of $p$ node pairs. This *pairwise scheme* runs in time $2^n (3/2)^p n^{O(1)}$ in space $2^n (3/4)^p n^{O(1)}$ for any $p = 0, 1, \ldots, n/2$. For instance, with $p = n/2$ we get time $O(2.4495^n)$ in space $O(1.733^n)$, thus improving upon the simple scheme at $s = 4/5n$ in both time and space (cf. the calculated bounds above).

So far we have assumed that the indegrees are at most some constant $k$. A straightforward implementation of the ideas described above yields time bounds where the number of possible parent *sets* appears as a multiplicative factor (absorbed into $n^{O(1)}$). Ideally, one would have that term affecting the running time only additively. The issue becomes very significant in practice when $k$ is large, and particularly so if the indegree is unbounded. With the Gurevich–Shelah recurrence we, unfortunately, have not found a way to reach this goal. In sharp contrast, our novel scheme is amenable to such an implementation. Specifically, we show in Section 5 that even if the maximum indegree $k$ is let to grow linearly with the number of nodes (with a decent slope), the indegree bound $k$ will play no role in the dominating part of the running time bound.

Finally, the pairwise scheme provides a desirable feature not possessed by previous exact algorithms: easy and efficient parallelization. We note that while the Silander–Myllymäki implementation is easy to run in parallel on $n$ processors, the time requirement per processor remains $2^n$, up to a polynomial factor (Silander and Myllymäki, 2006). The pairwise scheme goes beyond this. We show in Section 6 that the time requirement per processor decreases with $p$ as $2^n (3/4)^p$, thus yielding savings in both time and space, provided that sufficiently many ($2^p$) processors are available.

While the present work is mostly theoretical, we expect the techniques also to have practical significance. In Section 7 we present a preliminary empirical exploration of the potential of the pairwise scheme. With the current implementation we are not able to beat the record of 29 nodes, as parallelization remains to be implemented and certain key operations have not been optimized to the level of the Silander–Myllymäki implementation. However, we are currently working on a more streamlined implementation and expect networks of 34 nodes to be within reach, if employing some hundreds of processors in parallel.

## 2 PRELIMINARIES

In this section, we first formulate the problem of structure discovery in Bayesian networks. Then we present one of the possible variants of existing exact algorithms. Finally, we tune the problem formulation to accomodate for limited space.

### 2.1 The Structure Discovery Problem

A Bayesian network is a multivariate probability distribution that obeys a structural representation in terms of a directed acyclic graph (DAG) and a corresponding collection of univariate conditional probability distributions. For our purposes, it is crucial to treat the DAG, i.e., the network structure, explicitly, whereas the conditional probabilities will enter our formalism only implicitly.

A DAG on a set $N$ is an acyclic graph $(N, A)$ with node set $N$ and arc set $A$. A node $u$ is said to be a parent of $v$ if the arc $uv$ is in $A$. We denote by $A_v$ the set of parents of $v$. We associate the DAG with the edge set $A$ when there is no ambiguity about the node set. Throughout the paper we denote the cardinality of $N$ by $n$.

The problem of Bayesian network structure discovery is as follows. For each node $v \in V$ and a possible parent set $A_v \subseteq N \setminus \{v\}$ one specifies a local score $f_v(A_v)$ that gauges the fit of a class of conditional probability distributions to a given data set on the involved nodes under some statistical principle (e.g., Bayesian, maximum likelihood, minimum description length). Given these local scores, the task is to find a DAG $A$ that maximizes the sum of the local scores (Cooper and Herskovits, 1992; Heckerman et al., 1995),

$$f(A) \doteq \sum_{v \in N} f_v(A_v).$$

We note that this formulation does not directly apply to the so-called Bayesian approach to structure discovery (Friedman and Koller, 2003; Koivisto and Sood, 2004).

## 2.2 A Dynamic Programming Algorithm

While the existing exact algorithms for finding an optimal network structure exhibit some variability in the details, the key ideas are the same. We now review one of the variants, an algorithm the computes the maximum score—an actual DAG that achieves the optimal score can then be constructed using standard techniques. The algorithm can be described in two phases.

In the first phase, it computes for each node $v \in N$ and node subset $Y \subseteq N \setminus \{v\}$, the maximum of the local scores over the subsets of $Y$, defined as

$$\hat{f}_v(Y) \doteq \max_{X \subseteq Y} f_v(X). \tag{1}$$

In words, $\hat{f}_v(Y)$ is the maximum local score for $v$ given that the parents of $v$ must be selected from $Y$.

In the second phase, the algorithm effectively goes through all permutations of the nodes, however, tabulating only intermediate results for the *sets* of the first $i$ nodes in the order, for $i = 0, 1, \ldots, n$. Formally, we define $g(\emptyset) \doteq 0$ and for nonempty subsets $Y \subseteq N$ recursively:

$$g(Y) \doteq \max_{v \in Y} \left\{ g(Y \setminus \{v\}) + \hat{f}_v(Y \setminus \{v\}) \right\}. \tag{2}$$

In words, $g(Y)$ is the maximum score over DAGs on the node subset $Y$, obtained by selecting a node $v$ that, when being the last node in the order among the nodes in $Y$ and thus having the possibility to have parents from $Y \setminus \{v\}$, yields the largest score. In particular, $g(N)$ gives the maximum score over all DAGs on $N$.

The straighforward computation of the values $\hat{f}_v(Y)$ requires $2^{|Y|}$ steps for fixed $v$ and $Y$, and hence $n3^{n-1}$ steps in total. However, this can be significantly improved by dynamic programming, observed first in Ott and Miyano (2003):

**Lemma 1 (Ott and Miyano 2003)**

$$\hat{f}_v(Y) = \max \left\{ f_v(Y), \max_{u \in Y} \hat{f}_v(Y \setminus \{u\}) \right\}.$$

This recurrence can be solved in $O(n^2 2^n)$ steps.

Given the values $\hat{f}_v(Y)$, the recurrence for the values $g(Y)$ can then be computed in $O(n 2^n)$ steps in a straightforward manner. Thus, the algorithm takes a total of $O(n^2 2^n)$ steps, which is nearly optimal, since the input, as formulated above, may already consist of $n 2^{n-1}$ (real) numbers.

The space requirement of the above algorithm is $O(n 2^n)$ due to the space requirement of storing $\hat{f}_v(Y)$ for each $v$ and $Y$. This is possible to improve slightly to $O(\sqrt{n} 2^n)$ by noticing that the two phases can be merged into a single algorithm that visits every set $Y$ only once, proceeding level-wise, that is, in increasing cardinality of $Y$. This amounts to a truly significant saving in the space usage, provided that the input, the local scores $f_v(A_v)$, are given implicitly, that is, each score $f_v(A_v)$, once needed, is computed from (polynomial-space) input data; we next elaborate on this issue.

## 2.3 Limited Space: The Setting

To study the structure discovery problem with limited space usage, we need to be explicit about the input of the problem. To this end, it is convenient to let the local scores $f_v(A_v)$, for any node $v$, be available only for parent sets $A_v$ that belong to a given family of possible parent sets, denoted as $\mathcal{F}_v$; elsewhere we define $f_v(A_v) \doteq -\infty$. Whether the families $\mathcal{F}_v$ are represented implicitly or explicitly affects the space requirement of the structure discovery problem: In the former case, we assume each local score is evaluated based on input data once needed in time and space polynomial in $n$. In the latter case, the local scores are treated as explicit input, taking already space $\Omega(\sum_v |\mathcal{F}_v|)$. We will present our results under both approaches.

We are particularly interested in families $\mathcal{F}_v$ that are *downward closed*, that is, closed with respect to inclusion: if $Y \in \mathcal{F}_v$ and $X \subseteq Y$, then $X \in \mathcal{F}_v$. Important examples of such downward closed families are (a) the parent sets of cardinality at most $k$, for some fixed $k$, and (b) the parent sets that are contained in a given set of candidate parents (Perrier et al., 2008). A useful property of a downward closed family is that its members can be listed in essentially linear time. In fact, in Section 5 we will make use of the following slightly stronger observation; the proof (omitted) is, e.g., by visiting the members in increasing order of cardinality or in lexicographic order.

**Proposition 2** *Given a set $N$, a downward closed family $\mathcal{F} \subseteq 2^N$, and sets $X$ and $Y$ with $X \subseteq Y \subseteq N$, the members $Z \in \mathcal{F}$ with $X \subseteq Z \subseteq Y$ can be listed in time $|\mathcal{F}||N|^{O(1)}$.*

## 3　A DIVIDE & CONQUER SCHEME

We next consider in more detail the partitioning based approach outlined in the Introduction.

Let $\hat{A}$ be an optimal DAG on the node set $N$. Fix an integer $s$ with $n/2 \leq s \leq n$. Since $\hat{A}$ is acyclic there exists a partition of $N$ into two sets $N_0$ and $N_1$ of size $s$ and $n-s$, respectively, such that every arc between $N_0$ and $N_1$ in $\hat{A}$ is directed from $N_0$ to $N_1$; in other words, the parents of any node in $N_0$ are from $N_0$, while a node in $N_1$ may have parents also from $N_0$. That said, one can find $\hat{A}$—strictly speaking, the associated score $f(\hat{A})$—by trying out all possible partitions $\{N_0, N_1\}$ and solving the recurrences

$$g_0(Y) \doteq \max_{v \in Y} \left\{ g_0(Y \setminus \{v\}) + \hat{f}_v(Y \setminus \{v\}) \right\},$$

for $\emptyset \subset Y \subseteq N_0$ with $g_0(\emptyset) = 0$, and

$$g_1(Y) \doteq \max_{v \in Y} \left\{ g_1(Y \setminus \{v\}) + \hat{f}_v(N_0 \cup Y \setminus \{v\}) \right\},$$

for $\emptyset \subset Y \subseteq N_1$ with $g_1(\emptyset) = 0$; the score of $\hat{A}$ is obtained as the maximum of $g_0(N_0) + g_1(N_1)$ over all partitions $\{N_0, N_1\}$.

We notice that the two subproblems are independent of each other given the partition $\{N_0, N_1\}$, and thus can be solved separately. Applying the exact algorithm given in the previous section, the computation of $g_0$ takes time and space $2^s n^{O(1)}$. Computing $g_1$ can be more expensive, since evaluating the term $\hat{f}_v(N_0 \cup Y \setminus \{v\})$ requires the consideration of all the $2^s$ possible subsets of $N_0$ as parents of $v$, in addition to a subset from $Y \setminus \{v\}$. To simplify the analysis, we assume that the number of possible parent sets, $|\mathcal{F}_v|$, is polynomial in $n$, in which case $g_1$ can be computed in time and space $2^{n-s} n^{O(1)}$. Because there are $\binom{n}{s}$ possible partitions, we have the following result.

**Proposition 3** *The structure discovery problem in Bayesian networks can be solved in time $\binom{n}{s} 2^s n^{O(1)}$ in space $2^s n^{O(1)}$ for any $s = n/2, n/2+1, \ldots, n$, provided that each node has $n^{O(1)}$ possible, predetermined parent sets.*

The above approach yields a smooth time–space tradeoff for space bounds between $2^{n/2} n^{O(1)}$ and $2^n n^{O(1)}$. However, this is all but the end of the story: Firstly, within this range a more efficient scheme exists, as we will show in the next section. Secondly, with space less than $2^{n/2} n^{O(1)}$ the above scheme is not applicable, unless executed recursively, as we show next.

To solve the subproblems, namely computing $g_0(N_0)$ and $g_1(N_1)$, with less space we may apply the partitioning technique again. The problem of computing $g_0(N_0)$ being of the same form as the original problem, let us turn to look at the problem of computing $g_1(N_1)$. As before, we see that there exists a partitioning of the node set $N_1$ into subsets $N_{10}$ and $N_{11}$ such that every arc between $N_{10}$ and $N_{11}$ in the optimal DAG $\hat{A}$ is directed from $N_{10}$ to $N_{11}$. So, one can compute $g_1(N_1)$ by trying out all possible partitions $\{N_{10}, N_{11}\}$ and solving the recurrences

$$g_{10}(Y) \doteq \max_{v \in Y} \left\{ g_{10}(Y \setminus \{v\}) + \hat{f}_v((N_0 \cup Y \setminus \{v\})) \right\},$$

for $\emptyset \subset Y \subseteq N_{10}$ with $g_{10}(\emptyset) = 0$, and

$$g_{11}(Y) \doteq \max_{v \in Y} \left\{ g_{11}(Y \setminus \{v\}) + \hat{f}_v(N_0 \cup N_{10} \cup Y \setminus \{v\}) \right\},$$

for $\emptyset \subset Y \subseteq N_{11}$ with $g_{11}(\emptyset) = 0$; the score $g_1(N_1)$ is obtained as the maximum of $g_{10}(N_{10}) + g_{11}(N_{11})$ over all partitions $\{N_{10}, N_{11}\}$.

In general, one can apply partitioning recursively, say $d$ times, and then solve the remaining subproblems by dynamic programming. For an analysis of the time and space requirements, it is convenient to assume a balanced scheme: in every step of the recursion, the node set in question is partitioned into two sets of about equal sizes; for simplicity, assume $n$ is a power of 2. Then, at depth $d \geq 0$ of the recurrence, the node set in each subproblem in question is of size $s \doteq n/2^d$; hence, each subproblem can be solved in time and space $2^s n^{O(1)}$. Because each subproblem of size $2s$ is divided into $2\binom{2s}{s} \leq 2^{2s}$ subproblems of size $s$, the total number of subproblems of size $s$ that need to be solved is at most $2^n 2^{n/2} 2^{n/4} \cdots 2^{2s} = 2^{2n-2s}$.

**Theorem 4** *The structure discovery problem in Bayesian networks can be solved in time $2^{2n-s} n^{O(1)}$ in space $2^s n^{O(1)}$ for any $s = n/2, n/4, n/8, \ldots$, provided that each node has $n^{O(1)}$ possible, predetermined parent sets.*

As a theoretically interesting special case we have the following.

**Corollary 5** *The structure discovery problem in Bayesian networks can be solved in time $4^n n^{O(1)}$ in space polynomial in $n$, provided that each node has $n^{O(1)}$ possible, predetermined parent sets.*

## 4 PARTIAL ORDERS: THE PAIRWISE SCHEME

We next present a scheme for trading space against time when the exponential term in the space complexity ranges between $2^n$ and $2^{n/2}$.

In general terms, the idea is to fix a class of partial orders on the node set $N$ such that any linear order on $N$ can be realized as a linear extension of at least one of the partial orders in the class. Each partial order in the class corresponds to a restricted instance of the original task, which is to be solved by a suitably tailored variant of the basic dynamic programming algorithm. The bucket orders specified by the partitions of $N$ into two subsets of fixed sizes, described in the previous section, is a simple example of such a class. Below we introduce a more "efficient" class.

In what follows, we focus on partial orders that are specified by $p$ ordered pairs of nodes from $N$. More precisely, for a fixed integer $p$, with $0 \leq p \leq n/2$, we pick arbitrarily $2p$ distinct nodes $u_1, v_1, \ldots, u_p, v_p \in N$ and let $\mathcal{C}_p$ denote the set of all partial orders $\{\mu_1, \ldots, \mu_p\}$ such that each $\mu_q$ is either $u_q v_q$ (i.e., $u_q$ precedes $v_q$) or $v_q u_q$ (i.e., $v_q$ precedes $u_q$). Thus the cardinality of $\mathcal{C}_p$ is $2^p$. The following lemma tells us that the class $\mathcal{C}_p$ "covers" the linear orders on $N$; the proof is by picking up ordered pairs from a linear order.

**Lemma 6** *Let $p$ be an integer with $0 \leq p \leq n/2$, and let $\prec$ be a linear order on $N$. Then there exists a member $R \in \mathcal{C}_p$ such that $\prec$ is a linear extension of $R$, that is, $uv \in R$ implies $u \prec v$.*

Because of this coverage property an optimal DAG can be found by trying out every $R \in \mathcal{C}_p$ and searching for an optimal DAG compatible with $R$. A DAG $A$ is said to be *compatible* with a partial order $R$ if any topological ordering of $A$ is a linear extension of $R$.

In the dynamic programming algorithm (2) the restriction to linear extensions of $R$ amounts to simple constraints on the node subsets that need to be visited. We see that a set $Y \subseteq N$ can form the $|Y|$ first elements in a linear extension of $R$ if and only if $Y$ belongs to the family $\mathcal{N}_R$ defined as follows.

**Definition 1** *Let $R \in \mathcal{C}_p$. Denote by $\mathcal{N}_R$ the family of sets $Y \subseteq N$ satisfying*

$$v \in Y \quad \text{implies} \quad u \in Y \quad \text{for every } uv \in R.$$

Since $R \in \mathcal{C}_p$ contains $p$ disjoint pairs of nodes, and for each pair one of the four possibilities is excluded, we have the following.

**Lemma 7** *Let $R \in \mathcal{C}_p$. Then the cardinality of $\mathcal{N}_R$ is $3^p 2^{n-2p}$.*

The restricted dynamic programming algorithm evaluates the function $g^R$ defined by $g^R(\emptyset) \doteq 0$ and for nonempty $Y \in \mathcal{N}_R$ recursively:

$$g^R(Y) \doteq \max_{\substack{v \in Y \\ Y \setminus \{v\} \in \mathcal{N}_R}} \left\{ g^R(Y \setminus \{v\}) + \hat{f}_v(Y \setminus \{v\}) \right\}. \quad (3)$$

It follows that $g^R(N)$ equals the optimum score over DAGs compatible with $R$.

We summarize the above findings:

**Lemma 8** *Let $p$ be an integer with $0 \leq p \leq n$. Then for any partial order $R \in \mathcal{C}_p$ it holds that*

$$g^R(N) = \max\{f(A) : A \text{ is compatible with } R\},$$

*and, furthermore,*

$$\max_{R \in \mathcal{C}_p} g^R(N) = \max_A f(A),$$

*where $A$ runs through all DAGs on the $n$-node set $N$.*

We see that $g^R$ can be computed in time and space $|\mathcal{N}_R| n^{O(1)}$, provided that the number of possible parents for each node is polynomial in $n$. Thus we have obtained the following time–space tradeoff result, referred to as the *pairwise scheme*.

**Theorem 9** *The structure discovery problem in Bayesian networks can be solved in time $2^n (3/2)^p n^{O(1)}$ in space $2^n (3/4)^p n^{O(1)}$ for any $p = 0, 1, 2, \ldots, n/2$, provided that each node has $n^{O(1)}$ possible, predetermined parent sets.*

**Proof:** The space complexity is obtained by rewriting $|\mathcal{N}_R| = 3^p 2^{n-2p}$ as $2^n (3/4)^p$. The time complexity is obtained by multiplying $|\mathcal{N}_R| n^{O(1)} = 3^p 2^{n-2p} n^{O(1)}$ by $|\mathcal{C}_p| = 2^p$. □

The bounds in Theorem 9 should be compared to bounds that arise from the basic partitioning scheme (Proposition 3). For a comparison we set the space complexities equal (up to polynomial factors) and investigate the resulting running time bounds. To this end, we first solve $p$ from $2^n (3/4)^p = 2^s$, giving $p = (n-s)/\log_2(4/3)$. We also set $s = rn$, yielding the running time bound

$$2^{na(r)}, \quad a(r) \doteq r - r \log_2 r - (1-r) \log_2(1-r) \quad (4)$$

for the partitioning method and the bound

$$2^{nb(r)}, \quad b(r) \doteq 1 + (1-r) \log_2 \frac{2}{3} \Big/ \log_2 \frac{3}{4} \quad (5)$$

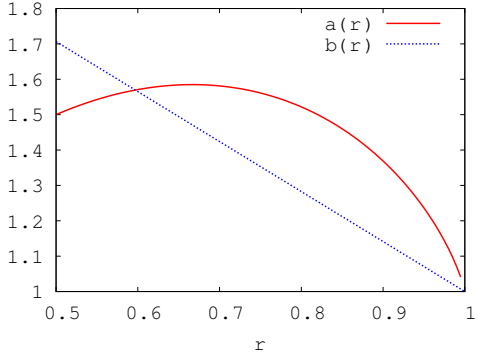

Figure 1: Comparison of the functions $a$ and $b$, defined in (4) and (5), respectively. Note that the values $b(r)$ correspond to the time complexity of the pairwise scheme only for $0.7924\ldots \leq r \leq 1$.

for the pairwise scheme. Note that the pairwise scheme only applies for $0 \leq p \leq n/2$, which in terms of $r$ amounts to $0.7924\ldots \leq r \leq 1$. Within this range, $b(r)$ appears to be strictly less than $a(r)$, except at $r=1$, in which case both schemes yield the running time bound of $2^n$ (up to polynomial factors); see Figure 1.

## 5 UNBOUNDED INDEGREE

So far we have assumed that the number of possible parent sets per node is polynomial in the number of nodes $n$. This is a reasonable assumption, e.g., when each node is allowed to have at most some constant number $k$ of parents. However, the polynomial factors in the running time bounds of Theorems 4 and 9 grow directly with the number of possible parent sets and thus heavily affect the performance in practice. In what follows, we aim at an implementation of the pairwise scheme (Theorem 9) such that the number of possible parent sets affects the running time *additively* rather than *multiplicatively*.

The idea is to arrange the computation of the recurrence (3) in such a way that some of the computations for $\hat{f}_v(Y \setminus \{v\})$, the maximum local score among the parent sets contained in $Y \setminus \{v\}$, are reused. Perhaps the most immediate attempt to do this would be dynamic programming according to the recurrence in Lemma 1; unfortunately, this seems to require space proportional to $2^n/\sqrt{n}$, for each $v$, when proceeding level-wise (Ott and Miyano, 2003): to compute the scores $\hat{f}_v(Y)$ for sets $Y$ of size $\ell$ one needs to access the scores $\hat{f}_v(X)$ of sets $X$ of size $\ell-1$. Luckily, it turns out that a sparse dynamic programming variant solves the task with less space.

Our key insight is the following simple observation, which we state in plain terms; the proof is omitted.

**Lemma 10** *Let $X$ and $Y$ be sets with $X \subseteq Y$. Let*

$$\mathcal{A} \doteq \{Z \subseteq Y : X \subseteq Z\},$$
$$\mathcal{B} \doteq \{Z \subseteq Y : x \notin Z \text{ for some } x \in X\}.$$

*Then $2^Y = \mathcal{A} \cup \mathcal{B}$ and $\mathcal{B} = \bigcup_{x \in X} 2^{Y \setminus \{x\}}$.*

In terms of the set functions $f_v$ and $\hat{f}_v$, for an arbitary $v$, the above lemma amounts to the following generalization of Lemma 1; the proof (omitted) is immediate by Lemma 10 and the definition of $\hat{f}_v$.

**Lemma 11** *Let $X$ and $Y$ be subsets of $N \setminus \{v\}$ with $X \subseteq Y$. Then*

$$\hat{f}_v(Y) = \max\{\max_{X \subseteq Z \subseteq Y} f_v(Z), \max_{u \in X} \hat{f}_v(Y \setminus \{u\})\}.$$

Note that Lemma 1 is obtained as a special case of Lemma 11, with $X = Y$.

Lemma 11 leaves us the freedom to choose a suitable node subset $X$ for each set of interest $Y$. How we make this choice is guided by the fact that, in the evaluation of $g^R$ by dynamic programming, we need the values $\hat{f}_v(Y)$ only for sets $Y$ that belong to $\mathcal{N}_R$; in what follows, we consider $R \in \mathcal{C}_p$ fixed. By storing the values $\hat{f}_v(Y)$ only for $Y \in \mathcal{N}_R$ we adhere to the space requirement (up to a polynomial factor) already needed for storing $g^R(Y)$ for each $Y \in \mathcal{N}_R$. Thus our goal is to choose $X$ such that $Y \setminus \{u\} \in \mathcal{N}_R$ for all $u \in X$. To this end, we let $X$ consist of all such nodes in $Y$ that have no larger node in $Y$ (w.r.t. $R$). Accordingly, for $Y \in \mathcal{N}_R$ define

$$X_Y \doteq \{u \in Y : uv \notin R \text{ for all } v \in Y\}.$$

Furthemore, define the *tail* of $Y$ as

$$\mathcal{T}_Y \doteq \{Z \subseteq Y : X_Y \subseteq Z\}.$$

In the next two lemmas we first show that $X_Y$ indeed has the desired property (in a maximal sense), and then that the tails for different sets $Y$ are pairwise disjoint, and thus optimally cover the subsets of $N$.

**Lemma 12** *Let $Y \in \mathcal{N}_R$ and $u \in Y$. Then $Y \setminus \{u\} \in \mathcal{N}_R$ if and only if $u \in X_Y$.*

**Proof:** "If": Let $u \in X_Y$. Let $st \in R$. By the definition of $\mathcal{N}_R$ we need to show that $t \in Y \setminus \{u\}$ implies $s \in Y \setminus \{u\}$. So, suppose $t \in Y \setminus \{u\}$, hence $t \in Y$. Now, since $Y \in \mathcal{N}_R$, we must have $s \in Y$. It remains to show that $s \neq u$. But this holds because $ut \notin R$ by the definition of $X_Y$.

"Only if": Let $u \notin X_Y$. Then we have $uv \in R$ for some $v \in Y$. But $u \notin Y \setminus \{u\}$ and $v \in Y \setminus \{u\}$, implying $Y \setminus \{u\} \notin \mathcal{N}_R$ by the definition of $\mathcal{N}_R$. □

**Lemma 13** *Let $Y$ and $Y'$ be two distinct sets in $\mathcal{N}_R$. Then the tails of $Y$ and $Y'$ are disjoint.*

**Proof:** Suppose the contrary that $Z \in \mathcal{T}_Y \cap \mathcal{T}_{Y'}$. W.l.o.g. let $w \in Y \setminus Y'$. Thus $w \notin Z$, for $Z \subseteq Y$ and $Z \subseteq Y'$. Therefore, $w \notin X_Y$ and $w \notin X_{Y'}$, for $X_Y \subseteq Z$ and $X_{Y'} \subseteq Z$. We conclude that $wv \in R$ for some $v \in Y$ and $wv' \in R$ for some $v' \in Y'$. But by the definition of $R$ we must have $v = v'$. Thus we have arrived at $wv \in R$, $w \notin Y'$, and $v \in Y'$, which is a contradiction given that $Y' \in \mathcal{N}_R$. □

We now merge the ingredients given above into an algorithm for evaluating $g^R$ using the recurrence (3), for a fixed $R \in \mathcal{C}_p$. In Algorithm 1 below, $g^R[Y]$ and $\hat{f}_v[Y]$ denote program variables that correspond to the respective target values $g^R(Y)$ and $\hat{f}_v(Y)$ to be computed. Also, recall that $\mathcal{F}_v$ denotes the family of possible parent sets for node $v$.

**Algorithm 1**

1. Let
$$g^R[\emptyset] \leftarrow 0\,.$$

2. For each $v \in N$, if $\emptyset \in \mathcal{F}_v$, then let
$$\hat{f}_v[\emptyset] \leftarrow f_v(\emptyset)\,; \quad \text{else let} \quad \hat{f}_v[\emptyset] \leftarrow -\infty\,.$$

3. For each nonempty $Y \in \mathcal{N}_R$, in increasing order of cardinality:

    (a) let
    $$g^R[Y] \leftarrow \max_{v \in X_Y} \left\{ g^R[Y \setminus \{v\}] + \hat{f}_v[Y \setminus \{v\}] \right\};$$

    (b) for each $v \in Y$ let $\hat{f}_v[Y]$ be the larger of
    $$\max_{Z \in \mathcal{T}_Y \cap \mathcal{F}_v} f_v(Z) \quad \text{and} \quad \max_{u \in X_Y} \hat{f}_v[Y \setminus \{u\}]\,.$$

**Lemma 14** *Algorithm 1 correctly computes $g^R$, that is, $g^R[Y] = g^R(Y)$ for all $Y \in \mathcal{N}_R$.*

**Proof:** By the definition of $g^R$ in (3) and by Lemma 11 it suffices to notice that, by Lemma 12, the condition "$v \in Y$ and $Y \setminus \{v\} \in \mathcal{N}_R$" is equivalent to $v \in X_Y$, given that $Y \in \mathcal{N}_R$. Note also that maximizing over $\mathcal{T}_Y \cap \mathcal{F}_v$ is equivalent to maximizing over $\mathcal{T}_Y$, since, by convention, $f_v(Z) = -\infty$ for $Z \notin \mathcal{F}_v$. □

We are ready prove the main result of this paper.

**Theorem 15** *The structure discovery problem in Bayesian networks can be solved in time $\left[2^n(3/2)^p + 2^pF\right]n^{O(1)}$ in space $2^n(3/4)^pn^{O(1)}$ for any $p = 0, 1, 2, \ldots, \lfloor n/2 \rfloor$, provided that for each node $v$ the family of possible parent sets $\mathcal{F}_v$ is downward closed and of size at most $F$.*

**Proof:** By Lemmas 8 and 14 it suffices to run Algorithm 1 for every $R \in \mathcal{C}_p$, that is, $|\mathcal{C}_p| = 2^p$ times.

The time requirement of Algorithm 1 is dominated by steps 3(a) and 3(b). Given $Y$, the set $X_Y$ can clearly be constructed in time $n^{O(1)}$. Thus the contribution of step 3(a) in the total time requirement is $|\mathcal{N}_R|n^{O(1)} = 3^p2^{n-2p}n^{O(1)}$ (Lemma 7).

We then analyze the time requirement of step 3(b), for fixed $v$. By Proposition 2 the maximization of the local scores over $\mathcal{T}_Y \cap \mathcal{F}_v$ can be done in time polynomial in $|\mathcal{T}_Y \cap \mathcal{F}_v|$. Since, by Lemma 13, these families are disjoint for different $Y \in \mathcal{N}_R$, the total contribution to the time requirement is polynomial in $|\mathcal{F}_v| \leq F$, for each $v$. Because step 3(b) is executed $|\mathcal{N}_R|$ times, the total time requirement of step 3(a) is $|\mathcal{N}_R|n^{O(1)} + Fn^{O(1)}$.

Combining the time bounds of step 3(a) and (b) and multiplying by $2^p$ yields the claimed bound $2^p\left[3^p2^{n-2p} + F\right]n^{O(1)} = \left[2^n(3/2)^p + 2^pF\right]n^{O(1)}$.

The space requirement is $|\mathcal{N}_R|n^{O(1)} = 3^p2^{n-2p}n^{O(1)}$, since, by Lemma 12, the values $g^R[Y]$ and $\hat{f}_v[Y]$ are needed only for $Y \in \mathcal{N}_R$. □

When there is no restrictions on the possible parent sets (i.e., each node has $2^{n-1}$ possible parent sets), we get the following.

**Corollary 16** *The structure discovery problem in Bayesian networks can be solved in time $2^{n+p}n^{O(1)}$ in space $2^n(3/4)^pn^{O(1)}$ for any $p = 0, 1, 2, \ldots, \lfloor n/2 \rfloor$.*

On the other hand, if each node is allowed to have at most $k$ parents, we get a significantly better running time bound, even if $k$ is let to grow linearly in the number of nodes $n$; the proof (omitted) is by direct comparison of the binomial coefficient $\binom{n-1}{k}$ and the bound $3^p2^{n-2p}$ in the stringest case, at $p = n/2$.

**Corollary 17** *The structure discovery problem in Bayesian networks can be solved in time $2^n(3/2)^pn^{O(1)}$ in space $2^n(3/4)^pn^{O(1)}$ for any $p = 0, 1, 2, \ldots, \lfloor n/2 \rfloor$, provided that each node is allowed to have at most $0.238n$ parents.*

## 6 ON PARALLELIZATION

We note that the pairwise scheme described in the previous sections allows for efficient parallelization. Obviously, each partial order $R \in \mathcal{C}_p$ can be treated in parallel. Furthermore, as in the Silander–Myllymäki implementation, the optimal local scores over given sets of possible parents in step 3(b) of Algorithm 1 can be precomputed—that is, not merging with step 3(a)—in parallel for each of the $n$ nodes. In total, this amounts to parallelization onto $2^p n$ processors (each with own memory); this is efficient in the sense that the running time per processor is scaled down by the same factor. So, if ignoring factors polynomial in $n$, the running time per processor becomes $2^n (3/4)^p$ (under the conditions of Corollary 17), thus exponentially less than $2^n$ when $p$ grows.

## 7 EMPIRICAL RESULTS

We have implemented the pairwise scheme in the C++ language. We examined the running time for the limit of 16 gigabytes of memory, letting the number of nodes $n$ vary from 25 to 31, with maximum indegree set to 3 (the local scores were taken as given, so computing them is not included in the running time estimates). First we estimated the minimum number of node pairs $p$ that yields a memory requirement of at most 16 gigabytes. Then we ran Algorithm 1 for a single partial order $R \in \mathcal{C}_p$; the resulting running time was multiplied by $2^p$ to get the total running time, see Table 7. The experiments were run on a 3.66-GHz Intel Xeon with 32 GB of RAM.

Table 1: The implemented pairwise scheme given 16 gigabytes of memory. Reported are CPU hours in total, $T$, and as divided to $2^p$ processors.

| $n$ | $p$ | $T$ | $T/2^p$ |
|---|---|---|---|
| 25 | 0 | 2 | 2.12 |
| 26 | 2 | 9 | 2.27 |
| 27 | 4 | 41 | 2.56 |
| 28 | 7 | 331 | 2.59 |
| 29 | 9 | 1660 | 3.24 |
| 30 | 12 | 13322 | 3.25 |
| 31 | 14 | 67748 | 4.14 |

We see that the current implementation is feasible up to around 31 nodes (4 weeks using 100 processors, 3 days using 1000 processors). However, we believe that by a more careful implementation both the time and the space requirement can be reduced to about one tenth, which should bring networks on 34 nodes to within reach (with massive parallelization).

**Acknowledgements**

The authors wish to thank Petteri Kaski, Fedor Fomin, Saket Saurabh, and Yngve Villanger for useful discussions on the Gurevich–Shelah recurrence. The research was supported in part by the Academy of Finland, Grant 125637.